\documentclass[runningheads,a4paper]{llncs}
\usepackage{amssymb}
\usepackage{graphicx,epsfig,setspace,subfig,url,amsmath}
\usepackage{algorithm}
\usepackage{algorithmic}
\usepackage{longtable}
\usepackage{array}
\usepackage{amsmath}
\usepackage{mathtools}
\usepackage{caption}

\urldef{\mailsb}\path|{gustavo.carneiro@adelaide.edu.au}|
\urldef{\mailsc}\path|{bradley@itee.uq.edu.au}|
\newcommand{\keywords}[1]{\par\addvspace\baselineskip\noindent\keywordname\enspace\ignorespaces#1}
 \normalsize

\begin{document}

\title{Automated 5-year Mortality Prediction using Deep Learning and Radiomics Features from Chest Computed Tomography}

\author{Gustavo Carneiro$^{\diamond}$ \qquad Luke Oakden-Rayner$^{\dagger}$ \qquad Andrew P. Bradley$^{\bullet}$  \qquad Jacinto Nascimentor$^{\ast}$  \qquad  Lyle Palmer$^{\dagger}$
\thanks{This work was partially supported by the Australian Research Council's Discovery Projects funding scheme (project DP140102794). Prof. Bradley is the recipient of an Australian Research Council Future Fellowship(FT110100623).}}
\institute {$^{\diamond}$ Australian Centre for Visual Technologies, The University of  Adelaide, Australia \\ 
$^{\dagger}$ School of Public Health, The University of Adelaide, Australia \\
$^{\bullet}$ School of ITEE, The University of Queensland, Australia \\
$^{\ast}$ Institute for Systems and Robotics, Instituto Superior T\'ecnico, Portugal}

\maketitle

\begin{abstract}
We propose new methods for the prediction of 5-year mortality in elderly individuals using chest computed tomography (CT).
The methods consist of a classifier that performs this prediction using a set of features extracted from the CT image and segmentation maps of multiple anatomic structures.
We explore two approaches: 1) a unified framework based on deep learning, where features and classifier are automatically learned in a single optimisation process; and 2) a multi-stage framework based on the design and selection/extraction of hand-crafted radiomics features, followed by the classifier learning process.
Experimental results, based on a dataset of 48 annotated chest CTs, show that the deep learning model produces a mean 5-year mortality prediction accuracy of 68.5\%, while radiomics produces a mean accuracy that varies between 56\% to 66\% (depending on the feature selection/extraction method and classifier). The successful development of the proposed models has the potential to make a profound impact in preventive and personalised healthcare.
\end{abstract}

\vspace{-.1in}
\section{Introduction}
\label{sec:intro}
\vspace{-.1in}

The prediction of reduced life expectancy in individuals is a public health priority and central to personalised medical decision making~\cite{ganna2015}.
Previous attempts to predict reduced life expectancy in the elderly have been studied using invasive (e.g., blood samples) and non-invasive (e.g., self-reported survey results, clinical examination) tests~\cite{ganna2015}.
These approaches resulted in a classification accuracy between 60\% and 80\%~\cite{ganna2015,yourman2012prognostic}, although patient age alone has shown a predictive accuracy of above 65\%~\cite{ganna2015}.
Compared to these previous attempts, the use of chest CT for the prediction of reduced life expectancy is advantageous because these scans potentially offer information on multiple organs and tissues from a single non-invasive test.  Hence, it is the aim of this paper to show that the use of chest CT alone (i.e., excluding previously used invasive and non-invasive tests) can produce accurate prediction of reduced life expectancy.

\begin{figure}[t]
\begin{center}
\includegraphics[width=3.8in]{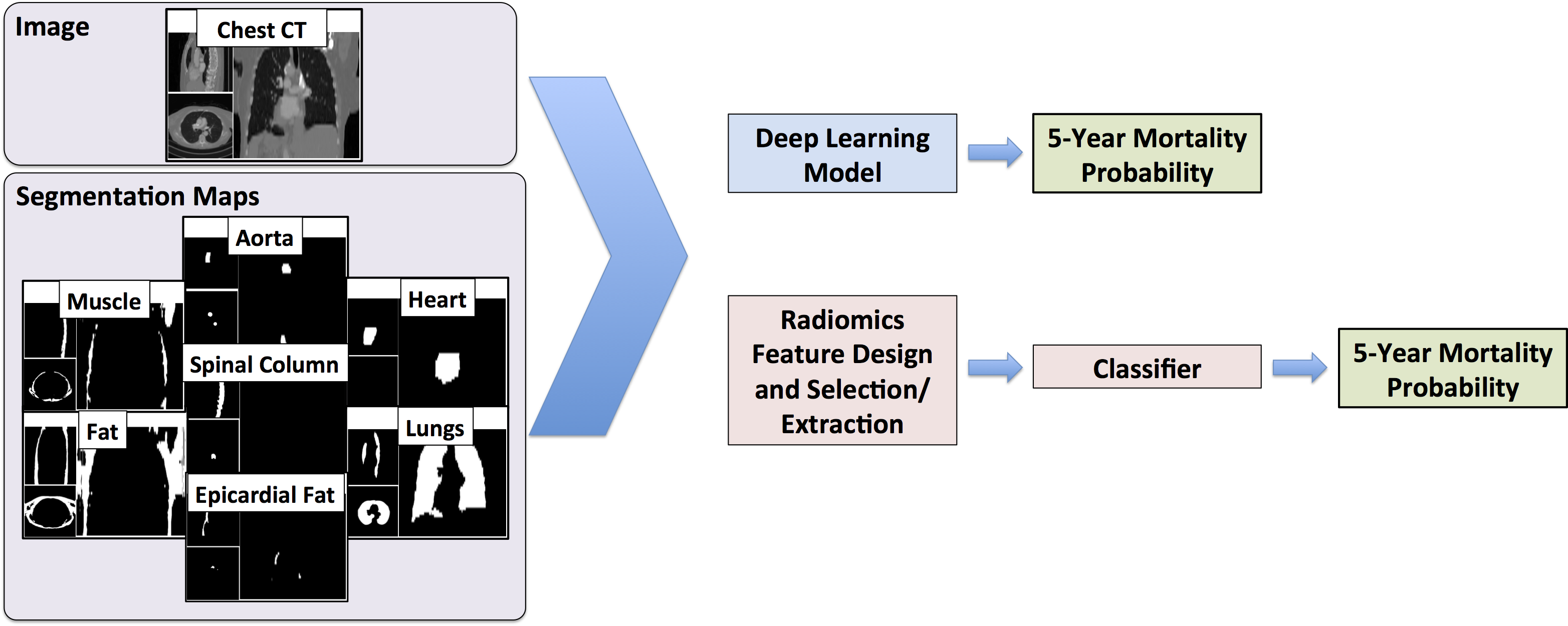} 
\end{center}
\caption{The proposed deep learning and radiomics models use image and segmentation maps to estimate the patient's 5-year mortality probability.}
\label{fig:main}
\end{figure}

Typically, prognostic models in medical image analysis have been designed for the prediction of disease specific outcomes~\cite{aerts2014decoding,lambin2012radiomics,kumar2012radiomics,bauer2007volumetric,haruna2010ct}, where the methodology requires hand-crafted features.
These features are selected/extracted based on their correlation with the prognosis, followed by modelling of the desired outcome using survival models or predictive classifiers.  
This multi-stage process of feature design and selection/extraction, followed by modelling has many disadvantages, such as the hand-crafting of the image features requiring medical expertise and being useful only for the particular prognosis being addressed, and the independence between feature selection/extraction and modelling potentially introducing redundant features and removing complementary features for the classification process.

In this paper, we propose two new approaches for the prediction of 5-year all-cause mortality in elderly individuals using chest CT and the segmentation maps of the following anatomies: aorta, spinal column, epicardial fat, body fat, heart, lungs and muscle. 
We have chosen chest CTs because they are commonly performed and widely available from hospitals, which facilitates dataset acquisition, and the segmentation maps are informed by previous biomarker research, which has demonstrated predictive and detectable changes in these tissues~\cite{kumar2012radiomics,bauer2007volumetric,haruna2010ct}.
The approaches developed in the paper are the following (Fig.~\ref{fig:main}): 1) a unified framework based on deep learning, where features and classifier are automatically learned in a single optimisation; and 2) a multi-stage framework based on the hand-crafting and selection/extraction of radiomics features, followed by a classifier learning process.
Experiments based on 48 annotated chest CT volumes show that the deep learning model produces mean classification accuracy of 68.5\%, while radiomics produces a mean accuracy that varies between 56\% to 66\% (depending on the feature selection/extraction method and classifier).  
Even though these results show comparable classification accuracy, deep learning models have an important  advantage compared to radiomics, which is the fully automated way of designing features, without requiring the assistance of a medical expert.  This advantage also means that future similar problems can be addressed in a more automated way, facilitating progress in this field of research.

\vspace{-.1in}
\section{Literature Review}
\label{sec:lit_review}
\vspace{-.1in}

This paper is related to radiomics and deep learning for medical image analysis.  Radiomics methods are a recent development in medical image analysis and are currently the state of the art in clinical studies. These methods are concerned with the design of hand-crafted features and their association with subtle variations in disease processes  (e.g. genetic variations)~\cite{lambin2012radiomics}.  Usually, radiomics methods are applied to imaging studies of patients with active tumours~\cite{aerts2014decoding}, but the application of these techniques to a general population of radiology patients for the prediction of important medical outcomes (e.g., mortality) is novel. 
The hand-crafting of features in these methods is inefficient because this process requires medical expertise, or alternatively if the features are task-agnostic (i.e. not informed by domain knowledge) it is not possible to know in advance which features will be effective, and it is therefore necessary to generate many possible features. 
This often requires a feature selection/extraction step to reduce the training complexity of the final classifier, and this step is based on a search heuristic that is not necessarily linked to the classification target.
For every new problem being addressed by radiomics, these two inefficient steps must be repeated, representing the major disadvantage of these methods.

Deep learning models are defined by a network composed of several layers of non-linear transformations that represent features of different levels of abstraction extracted directly from the input data~\cite{hinton2006reducing,krizhevsky2012imagenet}.  
In medical image analysis, deep learning can significantly improve segmentation and classification results~\cite{bar2015deep,ciresan2012deep,dhungel2015deep}, but its application to routinely collected medical images to predict important medical outcomes (e.g., mortality) has yet to be demonstrated. 
Our main references are the multi-view classification of mammograms~\cite{carneiro2015unregistered} the classifies breast exams into normal, benign and malignant; and the chest pathology classification using X-Rays~\cite{bar2015deep} because these works use deep learning methods for the high-level classification of medical images, but both classify diagnosis, which is conceptually different compared to our prognostic output.

\vspace{-.1in}
\section{Methodology}
\label{sec:methodology}
\vspace{-.1in}

\subsubsection{Dataset}
\label{sec:dataset}

The dataset is represented by ${\cal D} = \left \{ \left ( \mathbf{v}, \{ \mathbf{s}^{(j)} \}_{j \in \mathcal{A}}, y \right )_i \right \}_{i=1}^{|{\cal D}|}$, where ${\mathbf V}:\Omega \rightarrow \mathbb R$ denotes the chest CT with $\Omega \in \mathbb R^3$ representing the volume lattice of size ${w \times h \times d}$, $\mathbf{s}^{(j)} : \Omega \rightarrow \{0,+1\}$ represents the segmentation map for the anatomies in $\mathcal{A} = \{$ \emph{muscle, body fat, aorta, spinal column, epicardial fat, heart, and lungs} $\}$, and 
$y \in \{0,1\}$ denotes whether the patient is dead ($y=1$) or alive ($y=0$) on the time to censoring (time to death or time of last follow-up).

\vspace{-.1in}
\subsubsection{Radiomics}
\label{sec:radiomics}

This approach comprises the following stages~\cite{kumar2012radiomics}: 1) hand-crafting a large pool of features, 2) feature selection/extraction, and 3) classifier training.  
The hand-crafting process involves medical expertise to extract intensity, texture and shape information from particular image regions that are relevant for the final prognosis/diagnosis task.  
The feature extraction is denoted by
\begin{equation}
\mathbf{r} = r( \mathbf{v},\{ \mathbf{s}^{(j)} \}_{j \in \mathcal{A}} ),
\label{eq:radiomics_feature}
\end{equation}
where $r(.)$ represents a function that extracts the features $\mathbf{r} \in \mathbb{R}^R$.
Intensity features are based on the histogram of grey values $\mathbf{h}^{(j)} \in \mathbb{R}^H$ per anatomy $j \in \mathcal{A}$.
The feature is defined by statistics from $\mathbf{h}^{(j)}$, such as  mean, median, range, skewness, kurtosis, and etc.  In addition to these task-agnostic intensity-based features, we also include task-specific features that are related to the problem of estimating chronic disease burden, such as approximations of bone mineral density scoring (BMD)~\cite{bauer2007volumetric}, emphysema scoring~\cite{haruna2010ct}, and coronary (and aortic) artery calcification score~\cite{nasir2012interplay}.

The texture-based features use first and second-order matrix statistics, like the grey level co-occurrence matrix (GLCM) for anatomy $(j)$, denoted by $\mathbf{M}^{(j),d,a}_{GLCM}$, where the $r^{th}$ row and $c^{th}$ column of represent the number of times that grey levels $r$ and $c$ co-occur in two voxels
separated by the distance $d \in \mathbb{R}$ in the direction $a \in \mathbb{R}$ within the segmentation map provided by $\mathbf{s}^{(j)}$.  
The grey level run-length matrix (GLRLM) for anatomy $(j)$ is defined by $\mathbf{M}^{(j),a}_{GLRLM}$, where the $r^{th}$ row and $c^{th}$ column denote the number of times a run of length $c$ happens with grey level $r$ in direction $a$ within the segmentation $\mathbf{s}^{(j)}$. 
The grey level size-zone matrix (GLSZM) for anatomy $(j)$ is represented by $\mathbf{M}^{(j)}_{GLSZM}$, where the $r^{th}$ row and $c^{th}$ column denote the number of times  $c$ grey levels $r$ are contiguous in 8-connected pixels within the segmentation $\mathbf{s}^{(j)}$. 
Finally, the multiple gray level size-zone matrix (MGLSZM) for anatomy $(j)$ is defined by $\mathbf{M}^{(j)}_{MGLSZM}$, computed by a weighted average of several $\mathbf{M}^{(j)}_{GLSZM}$, each estimated with a different number of possible grey levels.
The features computed from these matrices are based on several statistics, such as energy, mean, entropy, variance, kurtosis, skewness, correlation, etc.  Each of the intensity and texture features are defined in a spatial context, by the use of weighted mean positions and spatial quartile means in all three dimensions, to identify any local variations across the tissues and organs.
Finally, the shape-based features are based on the volume of each anatomy $j \in \mathcal{A}$, computed from the segmentation map $\mathbf{s}^{(j)}$~\cite{kumar2012radiomics}.

The feature selection/extraction step forms a low-dimensionality vector $\widetilde{\mathbf{r}}  \in \mathbb{R}^{\widetilde{R}}$ ($\widetilde{R} << R$) using a heuristic that aims to reconstruct $\mathbf{r} \in \mathbb{R}^R$, under some constraints~\cite{tibshirani1996regression,jolliffe2002principal}. This vector is used for training the classifier, as in:
\begin{equation}
\gamma^* = \arg \min_{\gamma} \sum_{i \in \mathcal{T}} \Delta_{radiomics} \left ( y_i , g( \widetilde{\mathbf{r}}_i  ; \gamma ) \right ),  
\label{eq:classifier_radiomics}
\end{equation}
where $\mathcal{T} \in \mathcal{D}$ represents the training set, $g( \widetilde{\mathbf{r}}_i  ; \gamma  )$ denotes a classifier that returns a value in $[0,1]$ indicating the confidence in the 5-year mortality prediction, $\gamma$ represents the classifier parameters, and $\Delta_{radiomics}(.)$ denotes the loss function that penalises classification errors.

\vspace{-.1in}
\subsubsection{Deep Learning}
\label{sec:deeplearning}

The deep learning model used in this work is the Convolutional Neural Network (ConvNet)~\cite{lecun1995convolutional,krizhevsky2012imagenet}, defined as follows:
\begin{equation}
f( [\mathbf{v},\{ \mathbf{s}^{(j)} \}_{j \in \mathcal{A}}] ; \theta) = f_{out} \circ f_L \circ ... \circ f_2 \circ f_1([\mathbf{v},\{ \mathbf{s}^{(j)} \}_{j \in \mathcal{A}}] ; \theta_1),
\label{eq:CNN_1}
\end{equation}
where $\circ$ denotes the composition operator, $\theta$ represents the ConvNet parameters (i.e., weights and biases), and the output is a value in $[0,1]$ indicating the confidence in the 5-year mortality prediction.  Each network layer in (\ref{eq:CNN_1}) contains a set of filters, with each filter being defined by
\begin{equation}
\mathbf{x}(l+1) = f_l(\mathbf{x}(l); \theta_{l}) = \sigma(\mathbf{W}_{l}^{\top}\mathbf{x}(l) + \beta_{l}),
\label{eq:cnn_layer_definition}
\end{equation}
where $\sigma(.)$ represents a non-linearity~\cite{lecun1995convolutional}, 
$\mathbf{W}_l$ and $\beta_l$ denote the weight and bias parameters, and $\mathbf{x}(1) = [\mathbf{v},\{ \mathbf{s}^{(j)} \}_{j \in \mathcal{A}}]$. The last layer $L$ of the model in (\ref{eq:CNN_1}) produces a response $\mathbf{x}(L+1)$, which is the input for $f_{out}(.)$ that contains two output nodes (denoting the probability of 5-year mortality or survival), where layers $L$ and $out$ are fully-connected.
The training of the model in (\ref{eq:CNN_1}) minimises the binary cross entropy loss on the training set $\mathcal{T}$, as follows:
\begin{equation}
\theta^* = \arg \min_{\theta} \sum_{i \in \mathcal{T}} \Delta_{conv} \left ( y_i , f( \mathbf{x}_i(1)  ; \theta ) \right ),  
\label{eq:training_CNN}
\end{equation}
where $\Delta_{conv} \left ( y_i , f( \mathbf{x}_i(1)  ; \theta ) \right ) =  -y_i \times \log ( f( \mathbf{x}_i(1)  ; \theta  )) - (1-y_i) \times  \log (1 - f( \mathbf{x}_i(1)  ; \theta  ))$. 

\vspace{-.1in}
\section{Experiments}
\label{sec:experiment}
\vspace{-.1in}

\subsubsection{Materials and Methods}
\label{sec:materials}

The dataset has 24 cases (mortality) and 24 matched controls (survival), forming 48 annotated chest CTs of size $512 \times 512 \times 45$. Inclusion criteria for the mortality cases are: age $>$ 60, mortality in 2014, and underwent CT chest imaging in the 3 to 5 years preceding death. Exclusion criteria are: acute disease identified on CT chest, mortality unrelated to chronic disease (e.g., trauma), and active cancer diagnosis. 
Controls were matched on age, gender, time to censoring (death or end of follow-up), and source of imaging referral (emergency, inpatient or outpatient departments). 
Images were obtained using 3 types of scanners (GE Picker PQ 6000, Siemens AS plus, and Toshiba Aquilion 16) using standard protocols.
The chest CTs were obtained in the late arterial phase, following a 30 second delay after the administration of intravenous contrast (Omnipaque350/Ultravist370), and were annotated by a radiologist using semi-automated segmentation tools contained in the Vitrea software suite (Vital Images, Toshiba), where the following anatomies have been segmented: muscle, body fat, aorta, spinal column, epicardial fat, heart, and lungs.

The evaluation of the methodologies is based on a 6-fold cross-validation experiment, where each fold contains 20 cases and 20 matched controls for training and 4 cases and 4 matched controls for testing.
The classification performance is measured using the mean accuracy over the six experiments, with accuracy computed by $\frac{TP+TN}{TP + FP + TN + FN}$, where $TP$ represents correct mortality prediction, $TN$ denotes correct survival prediction, $FP$ means incorrect mortality prediction, and $FN$, incorrect survival prediction.
We also show the receiver operating characteristic (ROC) curve and area under curve (AUC)~\cite{hastie2005elements} using the classifier confidence on the 5-year mortality classification.

For the radiomics method, we hand-crafted 16210 features,where 2506 features come from the aorta, 2506 from heart , 2236 from lungs, 2182 from epicardial fat, 2182 from body fat, 2182 from muscle, and 2416 from spinal column~\footnote{Most of these features are hand-crafted with the methodology provided by J. Carlson (https://cran.r-project.org/web/packages/radiomics/).}, where 936 represent domain knowledge features~\cite{bauer2007volumetric,haruna2010ct,nasir2012interplay} (see Sec.~\ref{sec:methodology}).
For the feature selection/extraction, we have tried an identity linear feature extration (i.e., original features), LASSO~\cite{tibshirani1996regression} and PCA~\cite{jolliffe2002principal} learned with the training set for each fold. 
Finally, we tried different classifiers, such as linear (L) and non-linear (NL) support vector machine (SVM)~\cite{cortes1995support} and random forests (RF)~\cite{breiman2001random}.
Based on the experimental results, we show the performance of the following models: 1) features extracted with LASSO, NLSVM trained with $C=100$ and $\sigma = 0.01$;
2) original features, RF trained with with 900 trees, minimum \emph{nodesize} of 5 (minimum number of training samples per node), and trained with \emph{mtry} of 3 (i.e., number of variables sampled as candidates for each node split); and 3) 
features extracted with LASSO, LSVM trained with $C=100$.

The ConvNet has four convolutional layers, where the input has 8 channels (chest CT and 7 segmentation maps), the first layer has 50 filters and the second to fourth layers have 100 filters of size $5 \times 5 \times 2$ (i.e., these are 3-D filters).  The first convolutional layer has ReLU activation~\cite{nair2010rectified}, the fifth layer contains 6000 nodes, and the output layer has two nodes.  
For training, dropout~\cite{srivastava2014dropout} of $0.35$ is applied to all layers, the learning rate starts at $0.0005$, from epochs 1 to 10, which is then continuously reduced until it reaches $0.00001$ from epochs 60 to 120,  
and we use RMS prop~\cite{dauphin2015rmsprop} with $\rho=0.9$, and $\epsilon=10^{-6}$. This network and training parameters are selected based on their experimental results.

\vspace{-.1in}
\subsubsection{Results}
\label{sec:results}

\begin{figure}[t]
  \centering
    \includegraphics[width=1.0\textwidth]{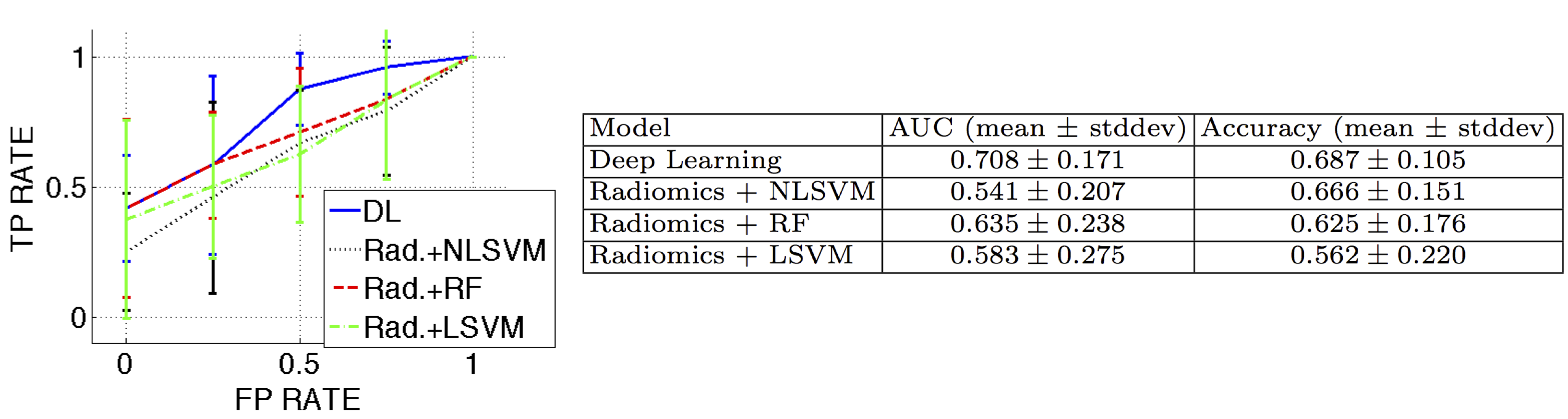} 
  \caption{Mean/standard deviation of the ROC (graph on left), AUC and accuracy (table on right) of the experiments on the testing set using the deep learning and radiomics with NLSVM, RF and LSVM classifiers.}
\label{fig:ROC}
\end{figure}

\begin{figure}[ht]
  \centering
  \begin{tabular}{cc}
    \includegraphics[width=0.27\textwidth]{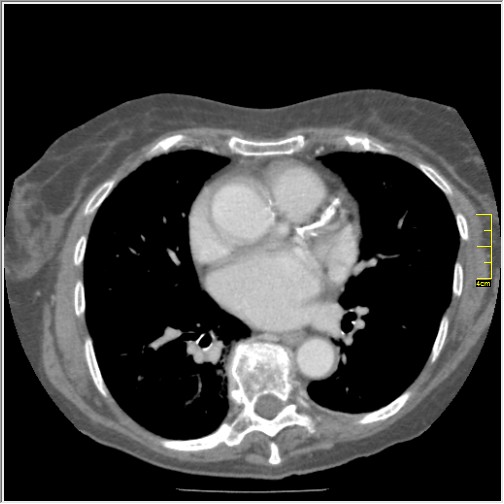} &
    \includegraphics[width=0.27\textwidth]{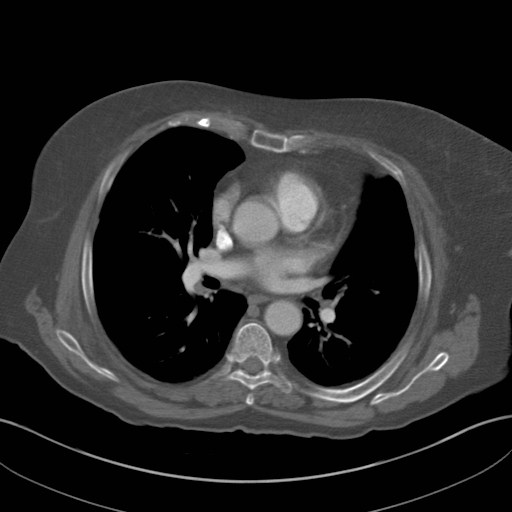} \\
	(a) Case (Mortality) & (b)  Control (Survival)  \\
DL: $f(.)=0.95$ & DL: $f(.)=0.42$\\
Rad.+NLSVM: $g(.)=0.55$ & Rad.+NLSVM: $g(.)=0.44$ \\
Rad.+RF: $g(.)=0.61$ & Rad.+RF: $g(.)=0.39$ \\
Rad.+LSVM: $g(.)=0.52$ & Rad.+LSVM: $g(.)=0.36$ \\
  \end{tabular}
  \caption{Testing examples of 5-year mortality classification produced by the deep learning (DL) and radiomics with NLSVM, RF, and LSVM with $f(.)$ in (\ref{eq:CNN_1}) and $g(.)$ in (\ref{eq:classifier_radiomics}) (both output the probability of 5-year mortality).  The example (a) shows heavy coronary artery calcification, enlarged aortic root and heart, low bone density, and moderate muscle mass loss.  Example (b) shows minimal artery calcification, mild muscle mass loss, preserved bone density and normal sized heart and aortic root.}
\label{fig:results}
\end{figure}

We show the mean and the standard deviation of the ROC curves for the testing set of the deep learning and the radiomics (with NLSVM, RF and LSVM classifiers) models in Fig.~\ref{fig:ROC}, which also shows a table with the mean and standard deviation of the AUC and accuracy of the testing set of the deep learning and the radiomics models.  
Using the t-test for paired samples, we note that there is no significant difference between any pair of models in terms of accuracy and AUC results on the testing set.
Also, all models are compared to the null hypothesis that the true mean is 0.5 (i.e., chance) for accuracy on the testing set, and both the deep learning and the radiomics with NLSVM classifier show a p-value $< 0.05$; for the AUC on the testing set, only the deep learning model shows a p-value $< 0.05$.
Finally, in Fig.~\ref{fig:results}, we show two chest CT examples with the output from both models.


\vspace{-.1in}
\section{Discussion and Conclusions}
\label{sec:discussion}
\vspace{-.1in}

The experiments demonstrate promising results, with prediction accuracy from routinely obtained chest CTs similar to the current state-of-the-art clinical risk scores, despite our small dataset and our exclusion of highly predictive covariates such as age and gender.  Furthermore, expert review of the correctly classified images (such as the example cases in Fig.~\ref{fig:results}) suggests that our models may be identifying medically plausible imaging biomarkers. The comparison between deep learning and radiomics models shows that they produce comparable classification results, but the deep learning model offers several advantages, such as automatic feature learning, and unified feature and classifier learning.

These advantages mitigate the issues of hand-crafting features, which requires expert domain knowledge, and the complicated multi-stage learning process of radiomics. It is in fact remarkable that a deep learning model implemented with relative simplicity could produce competitive results compared to the radiomics method, which uses features that have been heavily tuned for the task at hand~\cite{bauer2007volumetric,haruna2010ct,nasir2012interplay}, and relies on an extensive set of initial features (e.g., we have 16210 features). This hand-crafting task would need to be re-tuned for every new problem in radiomics, unlike the CNN approach. 
Finally, we believe that the deep learning results can be improved with the use of pre-training and data augmentation~\cite{hinton2006reducing,krizhevsky2012imagenet} and both models would benefit significantly from the integration of predictive epidemiological information (e.g., gender and age).

In this paper, we show the first proof of concept experiments for a system that is capable of predicting 5-year mortality in elderly individuals from chest CTs alone. 
The widespread use of medical imaging suggests that our methods will be clinically useful after being successfully tested in large scale problems (in fact, we are in the process of acquiring larger annotated datasets), as the only required inputs are already highly utilised: the medical images. We also note that the proposed deep learning model can be easily extended to other important medical outcomes, and other imaging modalities.

\vspace{-.15in}

\bibliographystyle{splncs}
\bibliography{miccai_ct_data.bib}
\end{document}